\RequirePackage{fix-cm}
\documentclass[twocolumn]{svjour3}          
\smartqed  
\usepackage{graphicx}
\usepackage{microtype}
\usepackage{hyperref}
\usepackage{rawfonts}
\usepackage{graphicx}
\usepackage{mathtools}
\usepackage{enumitem}
\usepackage{regexpatch}
\usepackage{multirow}
\usepackage{booktabs}
\usepackage{natbib}
\usepackage[utf8]{inputenc}
\usepackage[T1]{fontenc}
\usepackage{fancyvrb}

\usepackage{amsfonts}
\usepackage{todonotes}

\def\makeheadbox{\relax}

\newcommand{\setdef}[1]{\ensuremath{\{#1\}}}
\newcommand{\calP}{\mathcal P}
\newcommand{\Em}{\mathrm{Em}}
\newcommand{\Wb}{\mathrm{Wb}}
\newcommand{\Eq}{\mathrm{Eq}}
\newcommand{\Pref}{\mathrm{Pr}}

\DeclareMathOperator{\cpoEndOf} {\mathtt{endOf}}
\DeclareMathOperator{\cpoLengthOf} {\mathtt{lengthOf}}
\DeclareMathOperator{\cpoNoOverlap} {\mathtt{noOverlap}}
\DeclareMathOperator{\cpoEndBeforeStart} {\mathtt{endBeforeStart}}

\DeclareMathOperator{\cpoPresenceOf} {\mathtt{presenceOf}}
\DeclareMathOperator{\cpoAlternative} {\mathtt{alternative}}
\DeclareMathOperator{\cpoSpan} {\mathtt{span}}
\DeclareMathOperator{\cpoPulse} {\mathtt{pulse}}
\DeclareMathOperator{\cpoInterval} {\mathtt{interval}}
\DeclareMathOperator{\cpoSize} {\mathtt{size}}
\DeclareMathOperator{\cpoOptional} {\mathtt{optional}}
\DeclareMathOperator{\cpoStartOf} {\mathtt{startOf}}
\journalname{Journal of Scheduling}
\begin{document}

\title{Investigating Constraint Programming and Hybrid Methods for Real World Industrial Test Laboratory Scheduling
}


\author{Tobias Geibinger         \and
        Florian Mischek \and
        Nysret Musliu
}

\authorrunning{Tobias Geibinger, Florian Mischek, and Nysret Musliu} 

\institute{
Tobias Geibinger \at \email{tgeibing@dbai.tuwien.ac.at} \and
Florian Mischek \at \email{fmischek@dbai.tuwien.ac.at} \and 
Nysret Musliu \at \email{musliu@dbai.tuwien.ac.at} \\
\\
              Christian Doppler Laboratory for Artificial Intelligence and Optimization for Planning and Scheduling, Databases and Artificial Intelligence Group, TU Wien \\
              Favoritenstra\ss e 9-11, 1040 Vienna, Austria 
}

\date{Received: date / Accepted: date}

\maketitle

\begin{abstract}
In this paper we deal with a complex real world scheduling problem closely related to the well-known Resource-Constrained Project Scheduling Problem (RCPSP). The problem concerns industrial test laboratories in which a large number of tests has to be performed by qualified personnel using specialised equipment, while respecting deadlines and other constraints. We present different constraint programming models and search strategies for this problem. Furthermore, we propose a Very Large Neighborhood Search approach based on our CP methods. Our models are evaluated using CP solvers and a MIP solver both on real-world test laboratory data and on a set of generated instances of different sizes based on the real-world data.
Further, we compare the exact approaches with VLNS and a Simulated Annealing heuristic. We could find feasible solutions for all instances and several optimal solutions and we show that using VLNS we can improve upon the results of the other approaches. 
\end{abstract}

\section{Introduction}

Project scheduling includes various problems of high practical relevance. Such problems arise in many areas and include different constraints and objectives. Usually project scheduling problems require scheduling of a set of project activities over a period of time and assignment of resources to these activities. Typical constraints include time windows for activities, precedence constraints between them, assignment of appropriate resources etc. The aim is to find feasible schedules that optimize several criteria such as the minimization of total completion time.

In this paper we investigate solving a real-world project scheduling problem that arises in an industrial test laboratory of a large company.
This problem, Industrial Test Laboratory Scheduling (TLSP), which is an extension of the well known Resource-Constrained Project Scheduling Problem (RCPSP), was originally described by \cite{Mischek2021AnnalsOfOR,Mischek2018TechReport,MischekPatat18}. It consists of a grouping stage, where smaller activities (\emph{tasks}) are joined into larger \emph{jobs}, and a scheduling stage, where those jobs are scheduled and have resources assigned to them.
In this work, we deal with the second stage and assume that a grouping of tasks into jobs is already provided. Since we focus on the scheduling part, we denote the resulting problem TLSP-S.

A note on terminology: Different terms have been used in the literature to denote the units of work in project scheduling problems that should be scheduled: Jobs, tasks, projects, batches, and others.
In this paper we follow the convention by \cite{Brucker1999Survey} and use the term \emph{activities} when referencing general scheduling concepts or other project scheduling problems. Due to the specific distinction between tasks and jobs in TLSP(-S), we use these terms exclusively when referring to the corresponding TLSP(-S) concepts.

The investigated problem has several features of previous project scheduling problems in the literature, but also includes some specific features imposed by the real-world situation, which have rarely been studied before. 
Among others, these include heterogeneous resources, with availability restrictions on the activities each unit of a resource can perform. While work using similar restrictions exists \citep{DauzerePeres1998,Young17}, most problem formulations either assume homogeneous, identical units of each resource or introduce additional activity modes for each feasible assignment, which quickly becomes impractical for higher resource requirements and multiple resources.
Another specific feature of TLSP(-S) is that of linked activities, which require identical assignments on a subset of the resources. To the best of our knowledge, a similar concept appears only in \cite{Salewski1997ModeIdentity}, where modes should be identical over subsets of all activities.
We also deal with several non-standard objectives  which arise from various business objectives of our industrial partner. 
Most notably, we try to minimize the duration of each project, i.e. the time between the start of the first and the end of the last job in the project, instead of the usual makespan minimization, which is counted from the fixed release date of the project's first activity.

In practice, exact solutions for this problem are desired especially in situations where it is necessary to check if a feasible solution exists at all. In the application that we consider, checking quickly if activities of additional projects can be added on top of an existing schedule is very important. In this paper we investigate exact methods for solving this problem. Although it is known from previous papers \citep{Szeredi16,Young17} that constraint programming techniques can give good results for similar project scheduling problems, it is an interesting question if Constraint Programming (CP) techniques can also solve TLSP-S that includes additional features and larger instances. 

The current work is an extension of our previous conference paper \citep{Geibinger2019CPAIOR}.
In particular, we introduce a new Very Large Neighborhood Search (VLNS) heuristic based on the CP models and show that this hybrid approach leads to improvements in solution quality.
In addition, we provide a formal description of our CP model for the CP optimizer, and significantly extend our experimental section: We compare our results with those of a search heuristic that uses Simulated Annealing and apply our methods to a dataset taken from a real-life test laboratory.

The main contributions of this article are as follows:
\begin{itemize}
    \item We provide two different CP models for our problem by exploiting some previous ideas for a similar problem from \cite{Szeredi16} and \cite{Young17} and extending them to model the additional features of TLSP-S. This includes, for example, the handling of the problem specific differences discussed above but also new redundant constraints as well as search procedures tailored to the problem.
    Using the MiniZinc~\citep{Nethercote07} constraint programming language we experiment with various strategies involving the formulation of resource constraints, the reduction of the search space, and search procedures based on heuristics.
    The second model is geared towards the IBM CP Optimizer~\citep{cpoptimizer} and leverages several scheduling-specific features of that solver.
    \item We propose a new hybrid VLNS-based approach that internally makes use of the CP models.
    This heuristic iteratively fixes a part of the schedule and solves the resulting (smaller) instance to optimality.
    It also includes various strategies to compute lower bounds and avoid redundant computations or local optima.
    \item We evaluate our approaches both on real-world data from an industrial test laboratory and randomly generated instances that are based on the real-world data.
    Our experiments show that constraint programming techniques can reach good results for realistic instances and outperform MIP solvers on the same model. Our results strengthen the conclusion of previous studies and show that CP technology can be applied successfully for solving large project scheduling problems. 
    We also show that the hybridization of CP formulations with heuristics in the form of our VLNS approach is very successful in finding good solutions on instances of all sizes and further improves both upon the results of the CP models on the larger instances and those of a search heuristic based on Simulated Annealing.
\end{itemize}

The rest of the paper is organised as follows. In the next Section we give the related work. Section~\ref{sec:problem} introduces the problem that we investigate in this paper. A constraint model is given in Section~\ref{sec:model}, together with the alternative formulation for CP Optimizer. Section~\ref{sec:vlns} contains a description of the VLNS procedure. Experimental results are presented in Section~\ref{sec:results} and the last section gives conclusions.

\section{Literature Overview}
The Resource-Constrained Project Scheduling Problem (RCPSP) has been investigated by numerous researchers over the last decades.
For a comprehensive overview over publications dealing with this problem and its many variants, we refer to surveys e.g. by \cite{Brucker1999Survey}, \cite{Hartmann2010}, or \cite{Mika2015Survey}.

Of particular interest for the problem treated in this work are various extensions to the classical RCPSP.

Multi-Mode RCPSP (MRCPSP) formulations allow for activities that can be scheduled in one of several modes.
This variant has been extensively studied since 1977 \citep{Elmaghraby1977activity}, we refer to the surveys by \cite{Weglarz2011MRCPSPSurvey} and \cite{Hartmann2010}.
A good example of a CP-Model for the MRCPSP was given by \cite{Szeredi16}.

Many formulations, including TLSP, make use of release dates, due dates, deadlines, or combinations of those.
An example of this can be found in a paper by \cite{Drezet2008TimeConstraints}.
Further relevant extensions deal with multi-project formulations, including alternative objective functions \citep{Pritsker1969}. 
Usually, the objective in (variants of) RCPSP is the minimization of the total makespan \citep{Hartmann2010}.
However, also other objective values have been considered. 
Of particular relevance to TLSP are objectives based on project durations and multi-objective formulations (both appear in e.g. \cite{Nudtasomboon1997}).
\cite{Salewski1997ModeIdentity} include constraints that require several activities to be performed in the same mode.
This is similar to the concept of linked jobs introduced in the TLSP.

RCPSP itself and most variants assume that individual units of each resource are identical and interchangeable.
A problem closely related to TLSP-S is Multi-Skill RCPSP (MSPSP), first introduced by \cite{Bellenguez2005MSPSP}.
In this problem, each resource unit possesses certain skills, and an activity can only have those resources with the required skills assigned to it.
This is similar to the availability restrictions on resources that appear in TLSP.
Just like for our problem, they also deal with the problem that while availability restrictions could be modeled via additional activity modes corresponding to feasible resource assignments \citep{Bartels2009DestructiveMode,Pritsker1969,Schwindt2000Batch}, this is intractable due to the large number of modes that would have to be generated \citep{Bellenguez2005MSPSP}.
To the best of our knowledge, the best results for the MSPSP problem have been achieved by \cite{Young17}, who use a CP model to solve the problem.
We will be comparing and contrasting our CP model for TLSP-S with their model for MSPSP throughout this paper.

Also variants of VLNS have been used to solve the RCPSP or extensions of it, such as by \cite{Palpant2004}.
One of the main challenges in these approaches is the choice of a subset of activities that is selected for optimization at each step.
In TLSP-S, we have multiple projects over which most of the constraints are evaluated.
This makes single or multiple projects a natural choice for this subset, which we exploit in our heuristic.

\cite{Bartels2009DestructiveMode} describe a problem for scheduling tests of experimental vehicles. It contains several constraints that also appear in similar form in TLSP-S, but includes a different resource model and minimises the number of experimental vehicles used.

\section{Problem Description}
\label{sec:problem}
As mentioned before, we deal with a variant of TLSP \citep{Mischek2021AnnalsOfOR}, where we assume that a grouping of tasks into jobs is already provided for each project, and focus on the scheduling part of the problem instead (TLSP-S).
Thus, the goal is to find an assignment of a mode, time slot and resources to each (given) job, such that all constraints are fulfilled and the objective function is minimized.

In the following, we introduce the TLSP-S problem.

Each instance consists of a scheduling period of $h$ discrete \emph{time slots}.
Further, it lists resources of different kinds:
\begin{itemize}
	\item A set $E$ of \emph{Employees} (shorthand $\Em$) who are qualified for different types of jobs.
	\item A set $B$ of \emph{workbenches} ($\Wb$) with different facilities.
	\item Various auxiliary lab \emph{equipment} ($\Eq$) groups.
	Each group $g$ represents a set $G_g$ of similar devices.
	The set of all equipment groups is denoted $G^*$.
\end{itemize}

Further we have given the set of \emph{projects} labeled $P = \setdef{1,\dots,|P|}$, and the set of \emph{jobs} to be scheduled $J = \setdef{1,\dots,|J|}$.
For a project $p$, the jobs of this project are given as $J_p \subseteq J$.

Each job $j$ has several properties\footnote{In TLSP, these are derived from the tasks contained within a job. Since we assume the distribution of tasks into jobs to be fixed, they can be given directly as part of the input for TLSP-S.}:
\begin{itemize}
    \item A time window, given via a \emph{release date} $\alpha_j$ and a \emph{deadline} $\omega_j$. In addition, it has a \emph{due date} $\bar{\omega}_j$, which is similar to the deadline, except that exceeding it is only a soft constraint violation.
    \item A set of \emph{available modes} $M_j \subseteq M$, where $M$ is the set of all modes.
    \item A \emph{duration} $d_{mj}$ for each available mode $m \in M_j$.
    \item The resource requirements for the job $j$:
        \begin{itemize}
            \item The number of \emph{required employees} $r^\Em_{m}$ depends on the mode $m \in M_j$. 
            Each of these employees must be chosen from the set of \emph{qualified employees} $E_j \subseteq E$.
            Additionally, there is also a set of \emph{preferred employees} $E^\Pref_j \subseteq E_j$.
            \item The number of \emph{required workbenches} $r^\Wb_{j} \in \setdef{0,1}$. 
            If a workbench is required, it must be chosen from the \emph{available workbenches} $B_j \subseteq B$.
            \item For each equipment group $g \in G^*$, the job requires $r^\Eq_{gj}$ devices, which must be taken from the set of \emph{available devices} $G_{gj} \subseteq G_g$ for the group.
        \end{itemize}
    \item The \emph{predecessors} $\calP_j$ of the job, which must be completed before the job can start. 
    Precedence relations will only occur between jobs of the same project.
    \item \emph{Linked jobs} $L_j$ of this job. 
    All linked jobs must be performed by the same employee(s). 
    As before, such links only occur between jobs of the same project.
    \item Optionally, the job may contain \emph{initial assignments}.
        \begin{itemize}
            \item An \emph{initial mode} $\dot{m}_j$.
            \item An \emph{initial starting time slot} $\dot{s}_j$.
            \item \emph{Initial resource assignments}:
                For each employee $e \in E$, the boolean parameter $\dot{a}^\Em_{ej}$ indicates whether the employee $e$ is initially assigned to $j$.
                Analogously, $\dot{a}^\Wb_{bj}$ and $\dot{a}^\Eq_{dj}$ perform the same function for each workbench $b \in B$ and each device $d \in G_g, g \in G^*$, respectively.
        \end{itemize}
        Some or all of these assignments may be present for any given job.
\end{itemize}

Out of all jobs, a subset are \emph{started jobs} $J^\mathrm{S} \subseteq J$.
A started job will always fulfill the following conditions:
\begin{itemize}
    \item It must have a preassigned mode.
	\item Its start time must be set to 1.
	\item It must have initial resource assignments fulfilling all requirements.
\end{itemize}
The initial assignments of a started job must not be changed in the solution.


A complete description of all constraints of the original model can be found in \cite{Mischek2021AnnalsOfOR}. The hard and soft constraints that we consider for the TLSP-S will be described in the next section, where we will introduce the CP model.

The aim for this problem is to find an assignment of a mode, time slot and resources to each (given) job, such that all hard constraints are fulfilled and the violation of soft constraints is minimized.



\section{Constraint Programming Model}
\label{sec:model}
We developed our model using the solver-independent modeling language MiniZinc \citep{Nethercote07}. Using MiniZinc we can easily compare different solvers. Furthermore, previous studies have shown that CP gives very good results for similar project scheduling problems. Most notably, the approaches by \cite{Young17} and \cite{Szeredi16} for MSPSP and MRCPSP respectively. MiniZinc also enables the use of user defined search strategies, which were shown to be very effective for MSPSP \citep{Young17}. For both scheduling problems, the LCG solver Chuffed \citep{Chu11} was able to achieve very good results. 

In order to provide an additional comparison, we also modeled our problem with the IBM ILOG CP Optimizer~\citep{cpoptimizer}. This model uses a different formulation of constraints and decision variables than the MiniZinc model and is described in Subsection~\ref{subsec:cpo_model}. 

A solution for the scheduling problem is represented by the following decision variables: The start time variable $s_j$ assigns a start time to each job $j$. Similarly, for each job $j$, mode variable $m_j$ assigns it a mode. For resource assignments we need the following variables: For each job $j$, the variable $a^\Em_{e j}$ is set to 1 if employee $e$ is assigned to $j$ and 0 otherwise, the variable $a^\Wb_{b j}$ is 1 if $j$ is performed on workbench $b$ and 0 otherwise, and the variable $a^\Eq_{d j}$ is 1 if device $d$ is used by $j$ and 0 otherwise. This representation of the resource assignments will prove to be useful for modelling the unary resource requirements in Section~\ref{subsec:unary}.

\subsection{Basic Hard Constraints}
The following constraints follow directly from the problem definition. As usual, the logical connective ``and'' (i.e. conjunction) is denoted by $\land$ and $\rightarrow$ stands for implication.
\begin{align}
&\mathrlap{ s_j \geq \alpha_j \ \land \ (s_j + d_{m_j j}) \leq \omega_j} \qquad & j \in J \label{mzn:constraint_timewindow} \\
&s_j \geq (s_k + d_{m_k k}) \qquad & j \in J,\ k \in \calP_j \\
&m_j \in M_j \qquad & j \in J \\
&a^\Em_{e j} = 1 \ \rightarrow \ e \in E_j \qquad & j \in J,\ e \in E \\
&a^\Wb_{b j} = 1 \ \rightarrow \ b \in B_j \qquad & j \in J,\ b \in B \\
&a^\Eq_{d j} = 1 \ \rightarrow \ d \in G_{gj} \qquad & j \in J, g \in G^*,d \in G_g \\ 
&\sum_{e \in E} \ a^\Em_{e j} \ = \ r^\Em_{m_j}  \qquad & j \in J \\
&\sum_{b \in B} \ a^\Wb_{b j} \ = \ r^\Wb_j  \qquad & j \in J \\
&\sum_{d \in G_g} \ a^\Eq_{d j} \ = \ r^\Eq_{g j}  \qquad & j \in J, \ g \in G^* \\
&a^\Em_{e j} = a^\Em_{e k} \qquad & j \in J,\ k \in L_j,e \in E  \\
&s_j = 1  \qquad & j \in J^\mathrm{S}  \label{mzn:constraint_started}  
\end{align}
Constraint (1) makes sure that each job is executed in its time window, (2) enforces that the prerequisite jobs of a job are always completed before it starts. Constraints (3--6) ensure that assigned modes, employees, workbenches, and devices are available for the respective job. In order to make sure that each job has exactly as many resources as required, we have constraints (7--9). Furthermore, we need constraint (10) to make sure that linked jobs are assigned to the same employees and constraint (11) to fix the start time of jobs which are already started.

The above set of constraints is however not enough to ensure a valid solution. Additionally, we have to consider constraints which enforce that no resource (employee, workbench, or equipment) is assigned to two or more jobs at the same time. Like it was the case with MSPSP~\citep{Young17}, the constraints used for modeling those \emph{unary resource requirements} have a tremendous impact on the practicability of the model and in the next subsection we will present different options for modeling such constraints.

\subsection{Unary Resource Constraints}\label{subsec:unary}
We will now present three different approaches for modeling unary resource constraints, each of which is designed with CP solvers in mind. Two of those three quickly proved to be impractical for our problem.

\subsubsection{Time-indexed approach}
The probably most straightforward way to model the non-overuse of any resource at any given time is captured by the following constraints, where the variable $f_j = s_j + d_{m_j j}$ captures the end of job $j$.
\begin{align}
&\sum_{j \in J, s_j \leq t < f_j}  a^\Em_{e j} \ \leq  1  & e \in E, \ 1 \leq t \leq h \label{mzn:time_index1}\\
&\sum_{j \in J, s_j \leq t < f_j}  a^\Wb_{b j} \ \leq  1  & b \in B, \ 1 \leq t \leq h \label{mzn:time_index2}\\
&\sum_{j \in J, s_j \leq t < f_j}  a^\Eq_{d j} \ \leq  1  & \hspace{-1mm} g \in G^*, d \in G_g, 1 \leq t \leq h \label{mzn:time_index3}
\end{align}
\begin{figure*}[ht!]
\begin{center}
\begin{minipage}{0.9\textwidth}
\begin{Verbatim}[numbers=left]
constraint forall(e in Employees)(
  forall(t in TimeSlots)(
    sum(j in Jobs where startTime[j] <= t /\ t < (startTime[j] + duration[modeAssignment[j],j]))(
      employeeAssignment[e,j]
    ) <= 1
  )
);
\end{Verbatim}  
\end{minipage}
\end{center}
\caption{MiniZinc source code for Constraint (\ref{mzn:time_index1}).}\label{fig:mzn_example}
\end{figure*}
The number of constraints generated by MiniZinc based on (\ref{mzn:time_index1}--\ref{mzn:time_index3}) is of course directly dependent on the planning horizon $h$ and the total number of resources. To further illustrate this,  Figure~\ref{fig:mzn_example} shows the exemplatory source for the encoding of Constraint~\ref{mzn:time_index1} in MiniZinc. During compilation, MiniZinc has to unravel the all the quantifiers including the sum, which means the number of generated constraints is cubic.

Because of the long compilation time and the high computer resource consumption, it quickly became clear that for our larger instances the time-indexed approach is not efficient (The model did not compile within our experimental time out, even for the smallest instances).
This is of course not surprising since \cite{Young17} came to a similar conclusion for MSPSP. Hence, we discarded this option after some preliminary testing. 

\subsubsection{Overlap constraint}
For MSPSP, \cite{Young17} achieved their best results using a so-called \emph{order constraint}. This constraint basically enforces that two activities cannot overlap in their execution when they use a common resource. During the initial modeling phase we tried a very similar approach. First, we introduced the new predicate $\mathtt{overlap}$:
$$ \mathtt{overlap}(j,k) :=  s_k < (s_j + d_{m_j j}) \ \land \ (s_k + d_{m_k k}) > s_j $$
In MSPSP, resources are assigned to activities with respect to the needed skill of the activity. For the overlap constraint it is not important which skill requirement the resource contributes to, so \cite{Young17} had to introduce an auxiliary variable to express that a resource is used by an activity. We on the other hand assign the resources directly and thus can model our \emph{overlap constraint} without any new variables. 
\begin{align}
\mathtt{overlap}(j,k) \rightarrow ( \ & \bigwedge_{e \in E} (\neg a^\Em_{e j} \lor \neg a^\Em_{e k}) \;\; \land \nonumber \\
 &\bigwedge_{b \in B} (\neg a^\Wb_{b j} \lor \neg a^\Wb_{b k}) \;\;\; \land    \nonumber \\
 & \bigwedge_{g \in G^*, d \in G_g} (\neg a^\Eq_{d j} \lor \neg a^\Eq_{d k}) \; ) \nonumber \\
& \hspace{-8mm}j,k \in J, \ j \neq k, \ \alpha_k < \omega_j \land \omega_k > \alpha_j 
\end{align}
Just like with the time-indexed approach, it turned out that the overlap constraint produced too many constraints and was thus impractical for larger instances. This is interesting because \cite{Young17} had no such problems, but their biggest instances only had 60 resources and 42 activities, whereas we have instances with more than 300 resources and jobs, respectively. It should however be noted that \cite{Young17} reduced the number of generated constraints by considering only \emph{unrelated} activity pairs, i.e. activities which do not depend on the execution of each other via precedence constraints (related activities can obviously never overlap). We on the other hand generate constraints for all pairs of jobs which are allowed to overlap based on their release dates and deadlines. Comparing only unrelated jobs requires the computation of the transitive closure of the job precedence relation and because both our generated instances and the real-world instances have a lot of unrelated jobs, we don't expect any significant improvement.

\subsubsection{Cumulative constraints}
Another way to model the unary resource constraints is to use $\mathtt{disjunctive}$ or $\mathtt{cumulative}$ global constraints. We have chosen the latter for our model, since none of the solvers we use in our experiments directly implement $\mathtt{disjunctive}$. The $\mathtt{cumulative}$ constraint takes as input the start times, durations and resource requirements of a list of jobs and ensures that their resource assignments never exceed a given bound. This is of course a perfect way to enforce non overload of any resource and both MSPSP and MRCPSP have efficient models which make use of $\mathtt{cumulative}$ in some way~\citep{Szeredi16,Young17}. Formally, a cumulative constraint is of the form $\mathtt{cumulative}(S, D, R, b)$ where $S$, $D$, and $R$ are $n$-tuples of integer decision variables and $b$ is an integer constant. To illustrate the semantics of the constraint suppose we have $n$ activities with start times $S = \langle s_1, \dots , s_n \rangle$, durations $D = \langle d_1, \dots , d_n \rangle$, and resource usages $R = \langle r_1, \dots , r_n \rangle$. Let $\mathcal{A}_t = \{ i \in \mathbb{N}^+ \mid  i \leq n, \ s_i \leq t \leq s_i + d_i \}$ be the set of activities active at time point $t$. We say that the cumulative constraint above is satisfied, if for every time point $t$ it holds that $\sum_{i \in \mathcal{A}_t} r_i \leq b$.

In order to enforce the non-overload of any resource we need three constraints (one for each resource type).
\begin{align}
&\mathtt{cumulative}((s_j)_{j \in J}, \ (d_{m_j j})_{j \in J}, \ (a^\Em_{e j})_{j \in J}, \ 1) & \nonumber \\
& & \hspace{-50mm} e \in E \ \ \label{mzn:constraint_unary1} \\
&\mathtt{cumulative}((s_j)_{j \in J}, \ (d_{m_j j})_{j \in J}, \ (a^\Wb_{b j})_{j \in J}, \ 1) & \nonumber \\
& & \hspace{-50mm} b \in B \ \ \\
&\mathtt{cumulative}((s_j)_{j \in J}, \ (d_{m_j j})_{j \in J}, \ (a^\Eq_{d j})_{j \in J}, \ 1) & \nonumber \\
& & \hspace{-50mm} g \in G^*, d \in G_g \label{mzn:constraint_unary3} \ \ 
\end{align}
In contrast to our first two modeling approaches, this one turned out to scale well. Since the others performed so poorly on large instances, the rest of our experiments were performed with the $\mathtt{cumulative}$ unary resource constraints.

\subsection{Soft Constraints}
\label{subsec:soft_const}
There are several soft constraints in our problem definition~\citep{Mischek2018TechReport}. 

MiniZinc has no direct support for soft constraints, hence we define them as sums which should be minimised. Those sums are given as follows.

The first soft constraint depends on the number of jobs. 
Since we consider the job grouping fixed in TLSP-S, it is reduced to a constant value. 
In order to achieve comparability with results for TLSP, we keep it as a constant term $s_1 = w_1 \cdot |J|$ in the objective value.

We give preference to solutions in which the assigned employees of a job are taken from the set of preferred employees:
$$ s_2 = w_2 \cdot \sum_{j \in J} \sum_{e \in (E \setminus E^\Pref_j)} a^\Em_{e j} $$

For each project, the total number of employees assigned to it should be minimised:
$$ s_3 = w_3 \cdot \sum_{p \in P} \sum_{e \in E} \ ((\sum_{j \in J_p} a^\Em_{e j}) > 0) $$

For each job, violating its due date should be avoided:
$$ s_4 = w_4 \cdot \sum_{j \in J} \mathit{max}(s_j + d_{m_j j} - \bar{\omega}_j, 0) $$

Lastly, project durations should be as small as possible:
$$ s_5 = w_5 \cdot \sum_{p \in P} ( \mathit{max}_{j \in J_p}(s_j + d_{m_j j}) - \mathit{min}_{j \in J_p}(s_j) ) $$

The objective of the search is then given by $$\mathit{min} \ \sum_{2 \leq i \leq 5} s_i.$$

At the moment, the values of the weights $w_i$ ($1 \leq i \leq 5$) are being determined in correspondence with a real-world laboratory. Currently all these weights are set to $1$.

\subsection{Redundant Constraints}
\label{subsec:red_const}
While the formulation of the constraints given above correctly represents the problem, we also study redundant constraints to enhance solver performance. Such constraints allow CP solvers to propagate decisions in additional ways, potentially leading to conflicts earlier and thus pruning the search space.

Finding good redundant constraints for our problem proved to be very hard since the search space is usually very big and at the beginning of the search there is little knowledge about the final duration of the jobs. To deal with this issue we introduced a relaxed $\mathtt{cumulative}$ constraint enforcing a global resource bound.
\begin{align}
 \mathtt{cumulative}(&(s_j)_{j \in J}, \nonumber \\
 &(\mathit{min}_{m \in M_j}(d_{m j}))_{j \in J}, \nonumber \\
 &(\mathit{min}_{m \in M_j}(r^\Em_{m}) + r^\Wb_{j} + \sum_{g \in G^*} r^\Eq_{g j} \ )_{j \in J}, \nonumber \\
 &|E| + |B| + \sum_{g \in G^*}|G_g| \ ) \label{mzn:redundant1}
\end{align}
This enables the search to discard scheduling options which are impossible regardless of the chosen modes early on.

On top of that, we can also formulate more straightforward \texttt{cumulative} constraints which enforce the global resource bounds for each resource at any point in time.
\begin{alignat}{1}
 &\mathtt{cumulative}((s_j)_{j \in J}, (d_{m_j j})_{j \in J}, (r^\Em_{m_j})_{j \in J}, |E|) \label{mzn:redundant2.1} \\
 &\mathtt{cumulative}((s_j)_{j \in J}, (d_{m_j j})_{j \in J}, (r^\Wb_{j})_{j \in J}, |B|)  \label{mzn:redundant2.2}\\
 &\mathtt{cumulative}((s_j)_{j \in J}, (d_{m_j j})_{j \in J}, (r^\Eq_{g j})_{j \in J}, |G|) \ \ \   g \in G^* \label{mzn:redundant2.3}
\end{alignat}
Given the large search space, trying to restrict the scope of the decision variables seems like a worthwhile idea. We achieve this by using \emph{global cardinality constraints}. There are a number of global cardinality constraints in MiniZinc but we use ones of the form $\mathtt{gcc\_low\_up}(V, c, l, u)$ where $V = \langle v_1, \dots, v_n \rangle$ is an $n$-tuple of decision variables, $c$ is a domain element,  and $l$, $u$ are constant non-negative integers. Let $o = |\{ i \in \mathbb{N}^{+} \mid i \leq n, \ v_i = c \}|$ be the number of occurrences of the domain value $c$ in the assignment of variables $V$, then the constraint is satisfied if $l \leq o \leq u$.

The following constraints allow us to set tight bounds for the total number of resources which should be used. 
\begin{align}
\mathtt{gcc\_low\_up}(&(a^\Em_{e j})_{e \in E,j \in J}, \ 1, \nonumber \\
& \sum_{j \in J} \mathit{min}_{m \in M_j}(r^\Em_{m}),  \nonumber \\
&\sum_{j \in J} \ \mathit{max}_{m \in M_j}(r^\Em_{m})) \label{mzn:redundant3.1} \\
\mathtt{gcc\_low\_up}(&(a^\Wb_{b j})_{b \in B,j \in J}, \ 1,  \sum_{j \in J}  r^\Wb_j, \sum_{j \in J}  r^\Wb_j) \label{mzn:redundant3.2} \\
\mathtt{gcc\_low\_up}(&(a^\Eq_{d j})_{g \in G^*,d \in G_g,j \in J}, \ 1, \sum_{j \in J}\sum_{g \in G^*} \ r^\Eq_{g j},\nonumber \\
&\sum_{j \in J}\sum_{g \in G^*} \ r^\Eq_{g j}) & \label{mzn:redundant3.3}
\end{align}
Constraint (23) enforces that no more employees can be assigned than the sum of the highest possible employee requirements and no less than the sum of the minimum requirements. The other two constraints analogously ensure that the number of assigned workbenches and equipment is tightly bounded by the cumulative requirement of all jobs.

\subsection{Search Strategies}
\label{sec:search}
During initial testing results immediately showed that the default search strategy of Chuffed (or Gecode) was not even able to find feasible solutions for most instances. This was not surprising since \cite{Young17} already had a similar issue with MSPSP. However, they were able to improve their results drastically by employing a new MiniZinc search annotation called \texttt{priority\_search} which is supported by Chuffed~\citep{Feydy17}. In general, search annotations in MiniZinc allow us to influence the search by modifying the value or variable selection strategies of the underlying solver. The relatively new \texttt{priority\_search} enables the specification of more complex selection strategies which are not definable with other annotations. For example, it allows us to tell the solver that whenever a decision is to be made, then it should make assignments for the unassigned job with the smallest possible start time.

Based on the research of \cite{Young17}, we have experimented with four slightly different versions of $\mathtt{priority\_search}$:
\begin{enumerate}[label=(\roman*)]
    \item \texttt{ps\_startFirst\_aff}
    \item \texttt{ps\_startFirst\_ff}
    \item \texttt{ps\_modeFirst\_aff}
    \item \texttt{ps\_modeFirst\_ff}
\end{enumerate}
All four search strategies branch over the jobs and their resource assignments. The order of the branching is the same for all strategies and is determined by the smallest possible start times of the jobs in ascending order. For each branch, searches (i) and (ii) initially assign the smallest start time to the selected job followed by assigning it the mode which minimises the job duration. Search procedures (iii) and (iv) start with the mode assignment and then assign the start time. Once the start time and the mode have been assigned for the selected job, all of the search strategies make resource assignments for the job. Searches (i) and (iii) start by assigning those resources to the job which are available and have the biggest domain, whereas (ii) and (iv) start with assignments which are either unavailable or have only one value in their domain.  

\subsection{Alternative CP model}\label{subsec:cpo_model}
We also modeled our problem with CP Optimizer~\citep{cpoptimizer}.

In this model, all the decision variables are (possibly optional) intervals. Optional intervals are not necessarily present in a solution unless some constraints require it. Intervals which are not present are ignored by other constraints. Internally, intervals contain a start and an end time and when declaring an interval we can directly constrain their domains. The keywords $\mathtt{startOf}(a)$ and $\mathtt{endOf}(a)$ can be used to obtain the start and respectively end of an interval $a$ and $\mathtt{lengthOf}(a)$ returns its length. Furthermore, it is also possible to fix the size of an interval variable at declaration. Furthermore, the constraint $\mathtt{noOverlap}(I)$, where $I$ is a tuple of intervals, ensures that only the present intervals of $I$ do not overlap. The presence (or non-presence) of an interval $a$ can be enforced using the Boolean function $\mathtt{presenceOf}(a)$ which is true whenever $a$ is present and an appropriate constraint. The constraint $\mathtt{alternative}(a,I,n)$, where $a$ is an interval, $I$ a tuple of intervals and $n$ is a natural number, can also affect the presence of intervals. This constraint is true if whenever $a$ is present, also exactly $n$ intervals from $I$ are present as well. Further CP Optimizer constraints we use are $\mathtt{endBeforeStart}(a_1,a_2)$ enforcing that interval $a_1$ ends before the start of interval $a_2$, and $\mathtt{span}(a,I)$ which constrains the interval $a$ to start at the earliest start and end at the latest end of any interval in $I$.

For a complete formal definition of the CP Optimizer language and all the supported constraints, we refer to \cite{Laborie18}. 

In this model the decision variables are given by the following interval variables:
\vspace{-3mm}
\begin{align}
&\cpoInterval\ a_j \subset [\alpha_j, \omega_j) &  j \in J \nonumber\\
&\cpoInterval\ b_p &  p \in P \nonumber \\
&  \mathrlap{\ \! \cpoInterval\ a_{ij}\ \cpoOptional\ \cpoSize\ d_{ij}} \ &  j \in J, i \in M_j \nonumber \\
&\cpoInterval\ a^{\Em \mathrm{M}}_{eij}\ \cpoOptional\ &  j \in J, i \in M_j, e \in E_j \nonumber\\
&\cpoInterval\ a^{\Em}_{ej}\ \cpoOptional\ &  j \in J, e \in E_j \nonumber\\
&\cpoInterval\ a^{\Wb}_{bj}\ \cpoOptional\ &  j \in J, b \in B_j \nonumber\\
&\cpoInterval\ a^{\Eq}_{gdj}\ \cpoOptional\ &  j \in J, g \in G^*, d \in G_{gj} \nonumber
\end{align}
The intervals $a_j$ represents the jobs and are constrained to the time windows of the respective job. The second set of intervals $b_p$ are auxiliary variables representing the total duration of the projects. Those intervals enable an easy formulation of the project duration soft constraint. The next intervals $a_{ij}$ are optional and the presence of such an interval indicates that job $j$ is performed in mode $i$. For a job $j$ several employee allocations are possible depending on its mode. The presence of an optional interval $a^{EmM}_{eij}$ represents the allocation of employee $e$ to perform job $j$ in mode $i$. The last three sets of optional intervals are used to indicate resource allocation.

The hard constraints of the problem are encoded with the following constraints:
\begin{align}
& \cpoSpan(b_p, [a_j]_{j \in J_p}) & p \in P \hspace{-10mm} \label{eq5}\\
& \mathrlap{\cpoEndBeforeStart(a_k,a_j)} & k \in P_j \hspace{-10mm}\label{eq6}\\
& \mathrlap{\cpoAlternative(a_j, (a_{ij})_{ i \in M_j} , 1)} &  j \in J \hspace{-10mm}\label{eq7}\\
& \mathrlap{\cpoAlternative(a_{ij}, (a^{\Em \mathrm{M}}_{eij})_{e \in E_j} , r^{\Em}_{ij})} &  \nonumber \\
& &  j \in J,i \in M_j \hspace{-10mm}\label{eq8}\\
& \mathrlap{\cpoAlternative(a^{\Em}_{ej}, (a^{\Em \mathrm{M}}_{eij})_{i \in M_j} , 1)}  &  j \in J, e \in E_j \hspace{-10mm} \label{eq9} \\
&\mathrlap{\cpoNoOverlap((a^{\Em \mathrm{M}}_{eij})_{j \in J, i \in M_j : e \in E_j})} &   e \in E \hspace{-10mm} \label{eq10}\\
&\mathrlap{\cpoNoOverlap((a^{\Em}_{ej})_{j \in J : e \in E_j})} &   e \in E  \hspace{-10mm}\label{eq11}\\
&\mathrlap{\cpoAlternative(a_{j}, (a^{\Wb}_{bj})_{b \in B_j} , r^{\Wb}_{j})} &  j \in J \hspace{-10mm}\label{eq13}\\
&\mathrlap{\cpoNoOverlap((a^{\Wb}_{bj})_{j \in J : b \in B_j})} &  b \in B \hspace{-10mm} \label{eq14} \\
&\mathrlap{\cpoAlternative(a_{j}, (a^{\Eq}_{gdj})_{d \in G_{gj}} , r^{\Eq}_{gj})} & j \in J, g \in G^* \hspace{-10mm}\label{eq15} \\
&\mathrlap{\cpoNoOverlap((a^{\Eq}_{gdj})_{j \in J : d \in G_{gj}})} &  g \in G^*, d \in G_{g}  \hspace{-10mm}\label{eq16} \\
&\mathrlap{\neg \cpoPresenceOf(a_{ji})} & j \in J, i \in (M \setminus M_j) \hspace{-10mm} \\
&\mathrlap{\neg \cpoPresenceOf(a^{\Em}_{ej})} & j \in J, b \in (E \setminus E_j) \hspace{-10mm} \\
&\mathrlap{\neg \cpoPresenceOf(a^{\Wb}_{bj})} & j \in J, b \in (B \setminus B_j) \hspace{-10mm} \\
&\mathrlap{\neg \cpoPresenceOf(a^{\Eq}_{dj})} &  \nonumber \\
& & \hspace{-15mm} j \in J, g \in G^*,d \in (G_g \setminus G_{gj}) \hspace{-10mm} \\
&\mathrlap{\cpoPresenceOf(a^{\Em}_{ej})} & \nonumber \\
&\mathrlap{\quad=\cpoPresenceOf(a^{Em}_{ek})} & j \in J, k \in L_i, e \in E \hspace{-10mm} \label{eq12}\\
&\cpoStartOf(a_j) = 1  & j \in J^\mathrm{S} \hspace{-10mm}
\end{align}

Constraint (26) allows us to easily formulate the soft constraints which depend on the length of a project. The job precedences are enforced through Constraint (27). The alternative constraints (29--31) make sure that each job is assigned enough employees for the selected mode, while (31) and (32) express that no employee can be assigned to two jobs at the same time. The constraints (33--36) enforce the same for workbenches and equipment. In order to make sure that only available resources are assigned to the jobs, we need constraints (37--40). Lastly, linked jobs are modeled using (41) and started jobs with (42).  

We can also formulate $\mathtt{cumulative}$ constraints in CP Optimizer in the following way: $\sum_{i \leq n} \mathtt{pulse}(a_i,r_i) \leq b$, where $a_i,\dots, a_n$ are intervals, $r_i,\dots ,r_n$ their resource usages, and $b$ is the resource bound. The following redundant constraints are similar to the $\mathtt{cumulative}$ constraints defined in Section~\ref{subsec:red_const}.
\begin{alignat}{2}
& \sum_{j \in J} \sum_{m \in M_j} \ \cpoPulse(a_{jm},r^{\Em}_{jm}) \leq |E| \\
& \sum_{j \in J} \sum_{m \in M_j} \ \cpoPulse(a_{jm},r^{\Wb}_{jm}) \leq |B| \\
& \sum_{j \in J} \sum_{m \in M_j} \ \cpoPulse(a_{jm},r^{\Eq}_{jm}) \leq \sum_{g \in G^*}|G_g| 
\end{alignat}

Finally, the objective is to minimize the following formula which is just the sum of all soft constraints defined in Section~\ref{subsec:soft_const}.
\begin{alignat}{2}
&\mathit{min} \quad w_2 \cdot \sum \limits_{j \in J} \max(0,\cpoEndOf(a_j)-\overline{\omega}_j) & \nonumber \\
&\quad \quad + w_3 \cdot \sum_{j \in J} \sum_{e \in (E \setminus E^{Pr}_j)} \cpoPresenceOf(a^{\Em}_{ej}) & \nonumber \\
&\quad \quad + w_4 \cdot \sum_{p \in P} \sum_{e \in E} \big( 0 < \sum_{j \in J_p} \cpoPresenceOf(a^{\Em}_{ej}) \big) & \nonumber \\
&\quad \quad + w_5 \cdot \sum_{p \in P} \cpoLengthOf(b_p) & \nonumber 
\end{alignat}

\section{Very Large Neighborhood Search}
\label{sec:vlns}
Utilising our CP model we implemented a \emph{Very Large Neighborhood Search}. The basic idea is to start with a feasible but suboptimal solution and repeatedly fix most of the schedule except for a small number of projects and then try to find an optimal solution for the unfixed projects (each such intermediate optimization is a \emph{move}).

The basic steps of the search are the following:
\begin{enumerate}
    \item \emph{Find Initial Solution} \\
    In order for the VLNS to work, we need a feasible schedule for the instance. Since our CP model approach finds feasible solutions fast, we used it to provide an initial solution. In particular, we use Chuffed with our MiniZinc CP model, $\mathtt{priority\_search}$ and \emph{free search} disabled. This approach usually yields feasible schedules within one minute, even for large instances.
    
    \item \emph{Calculate lower bound for each project} \\
    In parallel to step one we use our CP approach to determine a lower bound for each project. This is done by solving each project once without regard for the other projects. The optimal objective is then a lower bound for the penalty of the project in the original problem. Since this can take a long time for bigger projects we define a runtime limit of 30 seconds for each project. If the limit is reached and no optimal solution could be found, we determine a heuristic lower bound for the project as follows: 
    We add up the number of jobs ($s_1$), the minimum number of different employees needed for the project ($s_3$), and the minimal duration of all jobs on the longest path in the job dependency graph ($s_5$). 
    For $s_2$ and $s_4$, 0 is used as the lower bound.
    
    \item \emph{Fix all but $k$ projects} \\
    Once we have an initial solution and the lower bounds we can start the actual heuristic. We start by selecting at random a combination of $k$ projects (initially, $k = 1$), with the following properties: All of them overlap in the current schedule (or, if there are no such combinations, have overlapping time windows) and at least one of the projects has potential for improvement i.e. the difference between the current penalty and the lower bound is bigger than zero. 
    
    The projects which are not in the selected combination are then fixed. To achieve this, we modify the time windows and availabilities of each job contained in a fixed project. The release and due dates are changed to correspond to the current start and end dates, the available modes and resources are changed to only include the current assignments, and finally the resource requirements are restricted to equal the assignments.
    
    The resulting instance is then further reduced by cutting away the fixed jobs outside of the merged time window of the selected projects. These removed jobs cannot be changed or influence the result and reducing the size of the instance improves compilation time.
    
    \item \emph{Perform move} \\
    After the preprocessing we try to find an optimal solution for the selected projects, where we again set a runtime limit of 30 seconds. The best assignment found within this limit is then applied to the current incumbent schedule if it does not increase the penalty of the solution. With a parameterised probability $hotStartProb$ we \emph{hot start} the CP solver. This means that the current assignment of the selected projects is given to the solver as an initial solution. The hot start functionality is based on work by \cite{Demirovic18} and has been integrated by us into the current version of the solver Chuffed. Hot starting the search however means that the search never accepts moves which change the schedule but do not decrease the penalty. This is why we only hot start with a given probability. If we do not hot start the solver, we additionally set up $\mathtt{priority\_search}$ to assign resources not in input order but randomly. This further enhances diversification.
    
    After we have performed the move, we save the selected combination of the $k$ projects in a list. Combinations contained in this list (or subsets of a combination in the list) are not selected again unless there has been a change in a job overlapping the contained projects.
    
    \item \emph{Possibly change $k$} \\
    If $k$ is bigger than 1 and the incumbent schedule has been changed by the last move, then we set $k$ back to 1. Otherwise, if there are no more eligible combinations with size $k$, we increase $k$ by one or -- with probability $jumpProb$ -- by two. If there no more combinations for any $k$, we terminate.
    
    After this we check if the current solution is equal to the sum of all project lower bounds. Should that be the case, then we have found an optimal solution and can terminate. If not, we go back to step 3 or terminate if we have reached the time limit of the solver.
    
\end{enumerate}

\section{Experiments and Comparison}
\label{sec:results}

 For our experimental evalution, we used MiniZinc 2.2.3~\citep{Nethercote07} with Chuffed 0.10.3~\citep{Chu11} and CPLEX 12.8.0~\citep{cplex}. Our VLNS which is described in Section~\ref{sec:vlns} was implemented in Java 8. Furthermore, we also experimented with the ILOG CP Optimizer 12.8.0~\citep{cpoptimizer} which was not run from MiniZinc but with the ILOG Java API. We have also tested Gecode~\citep{gecode} as an additional CP solver included in MiniZinc, but it quickly proved to be inferior to Chuffed on this model even when run with multiple threads. Regarding comparison to other approaches in the literature, to the best of our knowledge no solutions exist yet for the problem we consider in this paper.

The experiments in Sections \ref{sec:exp_search} -- \ref{sec:exp_comp} were ran on a benchmark server with 224GB RAM and two AMD Opteron 6272 Processors each with max. 2.1GHz and 8 physical cores. Since all of the solvers we experimented with are single threaded, we usually ran multiple independent sets of benchmarks in parallel.

For Section ~\ref{sec:exp_softconst} we employed a computational cluster with 13 nodes, each 
having 2 Intel Xeon CPUs E5-2650 v4 (max.~2.90GHz, 12 physical cores, no hyperthreading).

In addition to the solution approaches described in this paper, we also compare with a solution approach by \cite{Mischek2021AnnalsOfOR} using Simulated Annealing (SA).
SA (first described by \cite{Kirkpatrick1983SimulatedAnnealing}) is a well-known metaheuristic which works by repeatedly applying small random changes (moves) to a candidate solution.
Improvements are always accepted, while worsening moves are accepted with a certain probability that depends on the difference in the objective function and a parameter called \emph{temperature}. 
Higher temperatures lead to higher acceptance probabilities.
Over the course of the search, the temperature is slowly decreased.

\subsection{Instances}
We use a total of 30 randomly generated instances (based on real-life situations) of different sizes for our experiments. 
Those instances are listed in Table~\ref{tab:instances}.

While those instances were generated randomly, they are still modeled after real-world scenarios.
To this end, we developed an instance generator that can create instance of different sizes and properties, while matching the distribution of durations and resources of the real-world data.
The general idea is that a (feasible, but not necessarily optimal) reference solution is created first, which is then used as the basis for defining the properties of projects and jobs.
That way, it is always guaranteed that each instance has a feasible solution.
The details of how the instance generation works and what configurations are possible are described in \citep{Mischek2021AnnalsOfOR}.

For half of the instances the generator was configured to very closely match the real-world laboratory of our industrial partner (data set \emph{LabStructure}).
The other half of the instances (data set \emph{General}) is more general and makes full use of various problem features, such as the number and design of equipment groups, the structure of the dependency graph of each project and others.

The instances all have three modes: a \emph{single} mode requiring only one employee, a \emph{shift} mode which requires two employees but has a reduced duration, and an \emph{external} mode that requires no employees at all. In general, jobs can be done in single mode or optionally in shift mode. Some instances however also include jobs which can only be performed in external mode. Initial assignments appear only for jobs which are already started or are fixed to their current value via availability restrictions and time windows (this is the case for around 5\% of all jobs).

Since the benchmark instances were generated with the full TLSP in mind and in TLSP-S we take the initial job grouping as fixed and unchangeable, the instances had to be converted. This is achieved by viewing the jobs as the smallest planning unit and assigning the job parameters -- which are defined by the tasks contained in the job in TLSP -- directly to the jobs.

The 30 benchmark instances are a selection from a total of 120 generated instances. We chose the first two instances of each size (scheduling horizon and number of projects) and two additional instances for the 3 smallest sizes. 
This selection was necessary because of the long time it would have taken to experiment with all 120 instances.
In total, our benchmark data set contains instances which have 5 to 90 projects, between 7 and 401 jobs, and span 88 to 782 time slots.

In addition, we also include results for one real-life instance (Lab) taken directly from the lab of our industrial partner (in anonymized form).
This instance covers a scheduling period of over one year, at a granularity of two timeslots per working day.
It includes a reference schedule that is the manually created schedule actually used in the lab at the time the instance was created.

All instances can be downloaded from \url{https://www.dbai.tuwien.ac.at/staff/fmischek/TLSP/}.

\begin{table*}
\caption[The set of test instances used for the experiments]
		{The set of test instances used for the experiments.
		Shown are the data set the instance is taken from and the ID within that set.
		The following columns list the number of projects, jobs and the length of the scheduling period, followed by the number of employees, workbenches and equipment groups.
		The last columns contain the mean available modes, qualified employees and available workbenches per job, as well as the mean available devices per job and equipment group (only over those jobs that actually require at least one device of the group, about 10\% of all jobs).\\
		\textsuperscript{*}The discrepancy compared to the generated instances arises from the fact that several equipment groups were not yet considered for planning at the time this instance was created.}
		\centering
		\begin{tabular}{@{}rlcrrrrrrrrrr@{}}
			\toprule
			\textbf{\#} & \textbf{Data Set} & \textbf{ID} & \textbf{$|P|$} & \textbf{$|J|$} & \textbf{$h$} & \textbf{$|E|$} & \textbf{$|B|$} & \textbf{$|G^*|$} & \textbf{$\overline{|M_j|}$} & \textbf{$\overline{|E_j|}$} & \textbf{$\overline{|B_j|}$} & \textbf{$\overline{|G_{gj}|}$} \\ \midrule
            1 & General & 000 & 5 & 7 & 88 & 7 & 7 & 3 & 1.43 & 2.08 & 3.57 & 1.50 \\
            2 & General & 001 & 5 & 8 & 88 & 7 & 7 & 3 & 1.75 & 4.88 & 3.63 & 15.67 \\
            3 & LabStructure & 000 & 5 & 24 & 88 & 7 & 7 & 3 & 1.71 & 1.84 & 3.38 & 11.67 \\
            4 & LabStructure & 001 & 5 & 14 & 88 & 7 & 7 & 3 & 1.79 & 4.36 & 3.5 & 0.36 \\
            5 & General & 005 & 10 & 29 & 88 & 13 & 13 & 4 & 1.52 & 4.04 & 3.48 & 5.76 \\
            6 & General & 006 & 10 & 18 & 88 & 13 & 13 & 6 & 1.50 & 5.56 & 4.22 & 13.28 \\
            7 & LabStructure & 005 & 10 & 37 & 88 & 13 & 13 & 3 & 1.68 & 6.16 & 4.03 & 0.65 \\
            8 & LabStructure & 006 & 10 & 29 & 88 & 13 & 13 & 3 & 1.62 & 6.21 & 3.76 & 21.01 \\
            9 & General & 010 & 20 & 60 & 174 & 16 & 16 & 5 & 1.57 & 7.42 & 4.42 & 11.36 \\
            10 & General & 011 & 20 & 84 & 174 & 16 & 16 & 4 & 1.58 & 7.31 & 4.30 & 3.70 \\
            11 & LabStructure & 010 & 20 & 65 & 174 & 16 & 16 & 3 & 1.52 & 6.28 & 4.43 & 26.26 \\
            12 & LabStructure & 011 & 20 & 62 & 174 & 16 & 16 & 3 & 1.68 & 7.27 & 4.24 & 1.21 \\
            13 & General & 020 & 15 & 29 & 174 & 12 & 12 & 5 & 1.55 & 5.76 & 3.97 & 1.12 \\
            14 & LabStructure & 020 & 15 & 53 & 174 & 12 & 12 & 3 & 1.62 & 6.28 & 4.47 & 20.63 \\
            15 & General & 025 & 30 & 113 & 174 & 23 & 23 & 3 & 1.59 & 8.26 & 4.41 & 5.71 \\
            16 & LabStructure & 025 & 30 & 105 & 174 & 23 & 23 & 3 & 1.65 & 7.52 & 4.25 & 39.63 \\
            17 & General & 015 & 40 & 126 & 174 & 31 & 31 & 3 & 1.56 & 9.26 & 4.48 & 29.53 \\
            18 & LabStructure & 015 & 40 & 138 & 174 & 31 & 31 & 3 & 1.64 & 7.36 & 3.57 & 41.93 \\
            19 & General & 030 & 60 & 208 & 174 & 46 & 46 & 6 & 1.63 & 9.85 & 4.11 & 31.45 \\
            20 & LabStructure & 030 & 60 & 212 & 174 & 46 & 46 & 3 & 1.63 & 9.28 & 4.17 & 78.16 \\
            21 & General & 035 & 20 & 76 & 520 & 6 & 6 & 5 & 1.70 & 4.24 & 3.62 & 8.08 \\
            22 & LabStructure & 035 & 20 & 71 & 520 & 6 & 6 & 3 & 1.68 & 4.30 & 3.42 & 11.70 \\
            23 & General & 040 & 40 & 196 & 520 & 12 & 12 & 4 & 1.67 & 6.95 & 4.47 & 4.24 \\
            24 & LabStructure & 040 & 40 & 187 & 520 & 12 & 12 & 3 & 1.65 & 6.55 & 4.51 & 1.38 \\
            25 & General & 045 & 60 & 260 & 520 & 18 & 18 & 6 & 1.58 & 7.65 & 4.52 & 23.95 \\
            26 & LabStructure & 045 & 60 & 239 & 520 & 18 & 18 & 3 & 1.63 & 7.44 & 4.42 & 33.65 \\
            27 & General & 050 & 60 & 270 & 782 & 13 & 13 & 4 & 1.65 & 6.89 & 4.39 & 3.89 \\
            28 & LabStructure & 050 & 60 & 247 & 782 & 13 & 13 & 3 & 1.63 & 6.97 & 4.21 & 23.42 \\
            29 & General & 055 & 90 & 384 & 782 & 19 & 19 & 5 & 1.64 & 7.27 & 4.29 & 26.89 \\
            30 & LabStructure & 055 & 90 & 401 & 782 & 19 & 19 & 3 & 1.66 & 7.34 & 4.53 & 36.76 \\ \midrule
            Lab & - & - & 74 & 297 & 606 & 22 & 17 & 3 & 1.25 & 6.02 & 5.36 & 1\textsuperscript{*} \\ \bottomrule
		\end{tabular}
		
		\label{tab:instances}
	\end{table*}

\subsection{Search \& Redundant Constraints Experiments}
\label{sec:exp_search}
\begin{table*}
\caption{Priority Search \& Redundant Constraints Experiments (Runtime 30m)}
    \centering
    \begin{tabular}{l|l|r|r|r|r}
          \toprule
            \textbf{Search} & \textbf{Constraints} &  \textbf{\#sat} &  \textbf{\#opt} &  \textbf{\#best} &  \textbf{avg rel. dist. to best} \\
          \midrule

default & (\ref{mzn:constraint_timewindow}-\ref{mzn:constraint_started}), (\ref{mzn:constraint_unary1}-\ref{mzn:constraint_unary3}) & 12 & 8 & 8 & 11.43\% \\ 
default & (\ref{mzn:constraint_timewindow}-\ref{mzn:constraint_started}), (\ref{mzn:constraint_unary1}-\ref{mzn:constraint_unary3}), (\ref{mzn:redundant1}) & 13 & 8 & 8 & 14.1\% \\ 
default & (\ref{mzn:constraint_timewindow}-\ref{mzn:constraint_started}), (\ref{mzn:constraint_unary1}-\ref{mzn:constraint_unary3}), (\ref{mzn:redundant1}-\ref{mzn:redundant2.3}) & 13 & 8 & 8 & 14.68\% \\ 
default & (\ref{mzn:constraint_timewindow}-\ref{mzn:constraint_started}), (\ref{mzn:constraint_unary1}-\ref{mzn:constraint_unary3}), (\ref{mzn:redundant2.1}-\ref{mzn:redundant2.3}) & 13 & 8 & 8 & 14.04\% \\ 
default & (\ref{mzn:constraint_timewindow}-\ref{mzn:constraint_started}), (\ref{mzn:constraint_unary1}-\ref{mzn:constraint_unary3}), (\ref{mzn:redundant2.1}-\ref{mzn:redundant3.3}) & 13 & 8 & 8 & 13.25\% \\ 
default & (\ref{mzn:constraint_timewindow}-\ref{mzn:constraint_started}), (\ref{mzn:constraint_unary1}-\ref{mzn:constraint_unary3}), (\ref{mzn:redundant3.1}-\ref{mzn:redundant3.3}) & 12 & 8 & 8 & 12.99\% \\   
default & (\ref{mzn:constraint_timewindow}-\ref{mzn:constraint_started}), (\ref{mzn:constraint_unary1}-\ref{mzn:constraint_unary3}), (\ref{mzn:redundant1}-\ref{mzn:redundant3.3}) & 13 & 8 & 8 & 12.98\% \\ 
\texttt{ps\_startFirst\_aff} & (\ref{mzn:constraint_timewindow}-\ref{mzn:constraint_started}), (\ref{mzn:constraint_unary1}-\ref{mzn:constraint_unary3}) & 28 & \textbf{14} & 17 & 1.55\% \\ 
\texttt{ps\_startFirst\_aff} & (\ref{mzn:constraint_timewindow}-\ref{mzn:constraint_started}), (\ref{mzn:constraint_unary1}-\ref{mzn:constraint_unary3}), (\ref{mzn:redundant1}) & 28 & \textbf{14} & 17 & 1.54\% \\ 
\texttt{ps\_startFirst\_aff} & (\ref{mzn:constraint_timewindow}-\ref{mzn:constraint_started}), (\ref{mzn:constraint_unary1}-\ref{mzn:constraint_unary3}), (\ref{mzn:redundant1}-\ref{mzn:redundant2.3}) & \textbf{30} & \textbf{14} & 18 & 0.91\% \\ 
\texttt{ps\_startFirst\_aff} & (\ref{mzn:constraint_timewindow}-\ref{mzn:constraint_started}), (\ref{mzn:constraint_unary1}-\ref{mzn:constraint_unary3}), (\ref{mzn:redundant2.1}-\ref{mzn:redundant2.3}) & 28 & \textbf{14} & 17 & 1.55\% \\ 
\texttt{ps\_startFirst\_aff} & (\ref{mzn:constraint_timewindow}-\ref{mzn:constraint_started}), (\ref{mzn:constraint_unary1}-\ref{mzn:constraint_unary3}), (\ref{mzn:redundant2.1}-\ref{mzn:redundant3.3}) & \textbf{30} & \textbf{14} & \textbf{22} & \textbf{0.66\%} \\ 
\texttt{ps\_startFirst\_aff} & (\ref{mzn:constraint_timewindow}-\ref{mzn:constraint_started}), (\ref{mzn:constraint_unary1}-\ref{mzn:constraint_unary3}), (\ref{mzn:redundant3.1}-\ref{mzn:redundant3.3}) & 28 & \textbf{14} & 20 & 1.59\% \\ 
\texttt{ps\_startFirst\_aff} & (\ref{mzn:constraint_timewindow}-\ref{mzn:constraint_started}), (\ref{mzn:constraint_unary1}-\ref{mzn:constraint_unary3}), (\ref{mzn:redundant1}-\ref{mzn:redundant3.3}) & \textbf{30} & \textbf{14} & 16 & 0.78\% \\ 
\texttt{ps\_startFirst\_ff} & (\ref{mzn:constraint_timewindow}-\ref{mzn:constraint_started}), (\ref{mzn:constraint_unary1}-\ref{mzn:constraint_unary3}) & 28 & \textbf{14} & 18 & 1.16\% \\ 
\texttt{ps\_startFirst\_ff} & (\ref{mzn:constraint_timewindow}-\ref{mzn:constraint_started}), (\ref{mzn:constraint_unary1}-\ref{mzn:constraint_unary3}), (\ref{mzn:redundant1}) & 28 & \textbf{14} & 17 & 1.8\% \\ 
\texttt{ps\_startFirst\_ff} & (\ref{mzn:constraint_timewindow}-\ref{mzn:constraint_started}), (\ref{mzn:constraint_unary1}-\ref{mzn:constraint_unary3}), (\ref{mzn:redundant1}-\ref{mzn:redundant2.3}) & \textbf{30} & \textbf{14} & 17 & 0.95\% \\ 
\texttt{ps\_startFirst\_ff} & (\ref{mzn:constraint_timewindow}-\ref{mzn:constraint_started}), (\ref{mzn:constraint_unary1}-\ref{mzn:constraint_unary3}), (\ref{mzn:redundant2.1}-\ref{mzn:redundant2.3}) & \textbf{30} & \textbf{14} & 18 & 0.91\% \\ 
\texttt{ps\_startFirst\_ff} & (\ref{mzn:constraint_timewindow}-\ref{mzn:constraint_started}), (\ref{mzn:constraint_unary1}-\ref{mzn:constraint_unary3}), (\ref{mzn:redundant2.1}-\ref{mzn:redundant3.3}) & \textbf{30} & \textbf{14} & 21 & 0.75\% \\ 
\texttt{ps\_startFirst\_ff} & (\ref{mzn:constraint_timewindow}-\ref{mzn:constraint_started}), (\ref{mzn:constraint_unary1}-\ref{mzn:constraint_unary3}), (\ref{mzn:redundant3.1}-\ref{mzn:redundant3.3}) & 28 & \textbf{14} & 19 & 1.7\% \\ 
\texttt{ps\_startFirst\_ff} & (\ref{mzn:constraint_timewindow}-\ref{mzn:constraint_started}), (\ref{mzn:constraint_unary1}-\ref{mzn:constraint_unary3}), (\ref{mzn:redundant1}-\ref{mzn:redundant3.3}) & \textbf{30} & \textbf{14} & 16 & 0.86\% \\ 
\texttt{ps\_modeFirst\_aff} & (\ref{mzn:constraint_timewindow}-\ref{mzn:constraint_started}), (\ref{mzn:constraint_unary1}-\ref{mzn:constraint_unary3}) & 28 & \textbf{14} & 15 & 3.31\% \\ 
\texttt{ps\_modeFirst\_aff} & (\ref{mzn:constraint_timewindow}-\ref{mzn:constraint_started}), (\ref{mzn:constraint_unary1}-\ref{mzn:constraint_unary3}), (\ref{mzn:redundant1}) & 28 & \textbf{14} & 14 & 2.4\% \\ 
\texttt{ps\_modeFirst\_aff} & (\ref{mzn:constraint_timewindow}-\ref{mzn:constraint_started}), (\ref{mzn:constraint_unary1}-\ref{mzn:constraint_unary3}), (\ref{mzn:redundant1}-\ref{mzn:redundant2.3}) & \textbf{30} & \textbf{14} & 14 & 2.81\% \\ 
\texttt{ps\_modeFirst\_aff} & (\ref{mzn:constraint_timewindow}-\ref{mzn:constraint_started}), (\ref{mzn:constraint_unary1}-\ref{mzn:constraint_unary3}), (\ref{mzn:redundant2.1}-\ref{mzn:redundant2.3}) & \textbf{30} & \textbf{14} & 14 & 2.68\% \\ 
\texttt{ps\_modeFirst\_aff} & (\ref{mzn:constraint_timewindow}-\ref{mzn:constraint_started}), (\ref{mzn:constraint_unary1}-\ref{mzn:constraint_unary3}), (\ref{mzn:redundant2.1}-\ref{mzn:redundant3.3}) & \textbf{30} & \textbf{14} & 15 & 2.7\% \\ 
\texttt{ps\_modeFirst\_aff} & (\ref{mzn:constraint_timewindow}-\ref{mzn:constraint_started}), (\ref{mzn:constraint_unary1}-\ref{mzn:constraint_unary3}), (\ref{mzn:redundant3.1}-\ref{mzn:redundant3.3}) & 28 & \textbf{14} & 15 & 2.84\% \\ 
\texttt{ps\_modeFirst\_aff} & (\ref{mzn:constraint_timewindow}-\ref{mzn:constraint_started}), (\ref{mzn:constraint_unary1}-\ref{mzn:constraint_unary3}), (\ref{mzn:redundant1}-\ref{mzn:redundant3.3}) & \textbf{30} & \textbf{14} & 14 & 2.95\% \\ 
\texttt{ps\_modeFirst\_ff} & (\ref{mzn:constraint_timewindow}-\ref{mzn:constraint_started}), (\ref{mzn:constraint_unary1}-\ref{mzn:constraint_unary3}) & 28 & \textbf{14} & 15 & 3.32\% \\ 
\texttt{ps\_modeFirst\_ff} & (\ref{mzn:constraint_timewindow}-\ref{mzn:constraint_started}), (\ref{mzn:constraint_unary1}-\ref{mzn:constraint_unary3}), (\ref{mzn:redundant1}) & 28 & \textbf{14} & 14 & 2.41\% \\ 
\texttt{ps\_modeFirst\_ff} & (\ref{mzn:constraint_timewindow}-\ref{mzn:constraint_started}), (\ref{mzn:constraint_unary1}-\ref{mzn:constraint_unary3}), (\ref{mzn:redundant1}-\ref{mzn:redundant2.3}) & \textbf{30} & \textbf{14} & 14 & 2.75\% \\ 
\texttt{ps\_modeFirst\_ff} & (\ref{mzn:constraint_timewindow}-\ref{mzn:constraint_started}), (\ref{mzn:constraint_unary1}-\ref{mzn:constraint_unary3}), (\ref{mzn:redundant2.1}-\ref{mzn:redundant2.3}) & \textbf{30} & \textbf{14} & 14 & 2.68\% \\ 
\texttt{ps\_modeFirst\_ff} & (\ref{mzn:constraint_timewindow}-\ref{mzn:constraint_started}), (\ref{mzn:constraint_unary1}-\ref{mzn:constraint_unary3}), (\ref{mzn:redundant2.1}-\ref{mzn:redundant3.3}) & \textbf{30} & \textbf{14} & 15 & 2.68\% \\ 
\texttt{ps\_modeFirst\_ff} & (\ref{mzn:constraint_timewindow}-\ref{mzn:constraint_started}), (\ref{mzn:constraint_unary1}-\ref{mzn:constraint_unary3}), (\ref{mzn:redundant3.1}-\ref{mzn:redundant3.3}) & 28 & \textbf{14} & 16 & 2.83\% \\ 
\texttt{ps\_modeFirst\_ff} & (\ref{mzn:constraint_timewindow}-\ref{mzn:constraint_started}), (\ref{mzn:constraint_unary1}-\ref{mzn:constraint_unary3}), (\ref{mzn:redundant1}-\ref{mzn:redundant3.3}) & \textbf{30} & \textbf{14} & 14 & 2.95\% \\ 
          \bottomrule
    \end{tabular}
    \medskip
    
    \label{tab:ps_experiments}
\end{table*}

Table~\ref{tab:ps_experiments} shows the comparison of search procedures and redundant constraints described in Sections~\ref{sec:search} and \ref{subsec:red_const} respectively. The column \emph{\#sat} shows for how many instances (out of the 30 generated instances) the model-search combination found feasible solutions, in difference \emph{\#opt} contains the number of instances solved to optimality. Furthermore, the values in the column \emph{\#best} show the number of times the best solutions (in this set of experiments) was found for an instance, and \emph{avg rel. dist. to best} is the average relative distance to the best solution.

Each model was run using Chuffed with free search enabled. Free search alternates between user-defined and activity-based search on each restart. The time limit was set to 30 minutes for each instance. It can be easily seen that any version of \texttt{priority\_search} is vastly superior to the default search of Chuffed. \\ \texttt{priority\_search} strategies solve more instances to optimality with any of the redundant constraints and 15 of the configurations using \texttt{priority\_search} find a feasible solution for every instance. It should be noted that the fourteen optimally solved instances are the same over all configurations. They all have less than or equal to 20 projects. The search strategy \texttt{ps\_startFirst\_aff} in combination with redundant constraints (\ref{mzn:redundant2.1}-\ref{mzn:redundant3.3}) achieved the best results in general, which is why this strategy -- albeit slightly modified to support random value selection -- is also used in the VLNS. 

Which redundant constraints are important, seems to depend on the used search strategy. However, all \texttt{priority\_search} strategies achieve their best results using all redundant constraints but constraint (\ref{mzn:redundant1}). Furthermore, when comparing the configurations using no redundant constraints and the ones using only redundant constraint (\ref{mzn:redundant1}), there seems to be no significant improvement by adding the constraint except for the default search strategy. The effects of redundant constraints (\ref{mzn:redundant3.1})-(\ref{mzn:redundant3.3}) are also interesting. Adding them exclusively appears to have negative effects for all search configurations, whereas in conjunction with constraints (\ref{mzn:redundant2.1})-(\ref{mzn:redundant2.3}) they improve the results in almost all cases including the overall best configuration.

\subsection{VLNS Parameter Configuration}
As described in Section~\ref{sec:vlns}, we have 2 parameters for VLNS: There is the probability for hot starting a move called \emph{hotStartProb} and there is the probability that if we have to change the neighborhood -- i.e. the number of projects we modify in a single move -- we increase it not by 1 but by 2. The latter parameter is called \emph{jumpProb}.

We tested several different configurations of those parameters and one trend which was clearly distinguishable was that setting the \emph{hotStartProb} to zero gives rather bad results. An explanation for this is that not hot starting moves may cause them to timeout before they find an improvement which wastes a lot of runtime. For the other configurations there were no significant performance differences between them, but hot starting 80\% of the time and jumping 35\% performed slightly better than the rest. 

\begin{figure}
    \centering
    \includegraphics[width=0.45\textwidth]{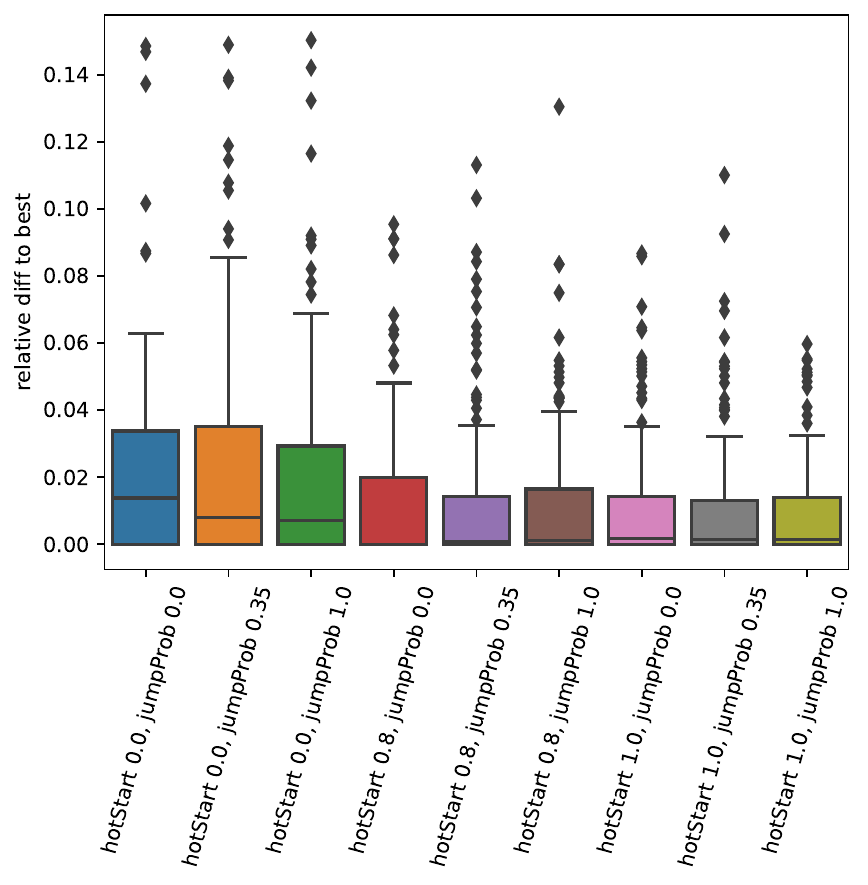}
    \caption{Comparison of selected VLNS configurations.}
    \label{fig:vlns}
\end{figure}

Figure~\ref{fig:vlns} shows the results of our experiments for selected configurations. We ran each instance with each configuration five times and each data point in the plot represents the relative difference of a run to the best solution found for the instance. It can be seen that regardless of the \emph{jumpProb}, the configurations where \emph{hotStartProb} is zero are clearly the worst. However, the other configurations show a very similar performance.

%

\subsection{Comparison}
\label{sec:exp_comp}
\begin{table*}
\caption{Comparison of different approaches with a timeout of two hours. Simulated Annealing and VLNS were run 5 times per instance with different seeds. Column \emph{Best} contains the best solution found, \emph{\#Feas.} the number of feasible solutions found (out of 5) and \emph{Avg} the average penalty over all feasible solutions.
			Solutions for VLNS, Chuffed, CPO and CPLEX marked with ``*'' indicate that the respective solver proved optimality.
			The best known solution for each instances is bold.}
			\centering
			\begin{tabular}{r|rcr|rlr|rl|rl|rl}
				\toprule
				\multicolumn{1}{c|}{\multirow{2}{*}{\textbf{\#}}} &
				\multicolumn{3}{c|}{\textbf{SA}} &
				\multicolumn{3}{c|}{\textbf{VLNS}} &
				\multicolumn{2}{c|}{\multirow{2}{*}{\textbf{Chuffed}}} &
				\multicolumn{2}{c|}{\multirow{2}{*}{\textbf{CPO}}} &
				\multicolumn{2}{c}{\multirow{2}{*}{\textbf{CPLEX}}} \\
				  & \multicolumn{1}{c}{\textbf{Best}} & \multicolumn{1}{c}{\textbf{\#Feas.}} & \multicolumn{1}{c|}{\textbf{Avg}} & 
				  \multicolumn{1}{c}{\textbf{Best}} & & \multicolumn{1}{c|}{\textbf{Avg}} & & & &         \\ \midrule
				1   & \textbf{98}   & 5    & 98     & \textbf{98}    & * & 98     & \textbf{98}   & * & \textbf{98}   & * & \textbf{98}   & * \\
				2   & \textbf{73}   & 5    & 73     & \textbf{73}    & * & 73     & \textbf{73}   & * & \textbf{73}   & * & \textbf{73}   &  \\
				3   & 152  & 5    & 152.6  & \textbf{149}   & * & 149    & \textbf{149}  & * & \textbf{149}  & * & \textbf{149}   &  \\
				4   & 106  & 5    & 107    & \textbf{105}   & * & 105    & \textbf{105}  & * & \textbf{105}  & * & \textbf{105}   & * \\
				5   & 287  & 5    & 306.4  & \textbf{283}   &   & 283    & \textbf{283}  & * & \textbf{283}  &   & 287   &  \\
				6   & 180  & 5    & 181.4  & \textbf{162}   & * & 162    & \textbf{162}  & * & \textbf{162}  &   & \textbf{162}   &  \\
				7   & \textbf{307}  & 5    & 307    & \textbf{307}   &   & 307    & \textbf{307}  & * & \textbf{307}  &   & --   &  \\
				8   & \textbf{310}  & 5    & 310    & \textbf{310}   &   & 310    & \textbf{310}  & * & \textbf{310}  &   & 317   &  \\
				9   & \textbf{501}  & 5    & 501    & \textbf{501}   & * & 501    & \textbf{501}  & * & 502  &   & --   &  \\
				10  & \textbf{564}  & 5    & 564    & \textbf{564}   & * & 564    & 576  &   & 566  &   & --   &  \\
				11  & 872  & 5    & 872.8  & \textbf{856}   &   & 856.4  & \textbf{856}  & * & 859  &   & --   &  \\
				12  & 660  & 5    & 660    & \textbf{656}   & * & 656    & \textbf{656}  & * & 661  &   & --   &  \\
				13  & 352  & 5    & 377.4  & \textbf{340}   & * & 340    & \textbf{340}  & * & \textbf{340}  &   & 370   &  \\
				14  & 423  & 5    & 424.2  & \textbf{420}   & * & 420    & \textbf{420}  & * & 424  &   & --   &  \\
				15  & 1085 & 5    & 1086.8 & \textbf{1084}  & * & 1085   & 1646 &   & 1104 &   & --   &  \\
				16  & 1141 & 5    & 1142   & \textbf{1138}  & * & 1142.6 & 1560 &   & 1193 &   & --   &  \\
				17  & 1195 & 5    & 1200.2 & \textbf{1194}  & * & 1194.4 & 1254 &   & 1203 &   & 1242   &  \\
				18  & \textbf{1359} & 5    & 1362.2 & \textbf{1359}  &   & 1378.2 & 1817 &   & 1404 &   & --   &  \\
				19  & 2127 & 5    & 2168.6 & \textbf{1964}  &   & 2000.8 & 2650 &   & 2099 &   & --   &  \\
				20  & 2228 & 5    & 2242.2 & \textbf{2114}  &   & 2140.8 & 2887 &   & 2284 &   & --   &  \\
				21  & 685  & 5    & 689    & \textbf{679}   & * & 679.2  & \textbf{679}  & * & 722  &   & --  &  \\
				22  & 766  & 5    & 768.4  & \textbf{765}   &   & 765    & \textbf{765}  & * & 770  &   & --   &  \\
				23  & 2209 & 4    & 2319.25& \textbf{2029}  &   & 2104.2 & 3486 &   & 2207 &   & --   &  \\
				24  & 1780 & 5    & 1817.2 & \textbf{1776}  &   & 1798.6 & 2445 &   & 1863 &   & --   &  \\
				25  & 2701 & 5    & 2836   & \textbf{2112}  &   & 2165.8 & 3278 &   & 2551 &   & --   &  \\
				26  & 2579 & 5    & 2599.8 & \textbf{2558}  &   & 2573.8 & 3896 &   & 2799 &   & --   &  \\
				27  & 2153 & 5    & 2155.2 & \textbf{2151}  & * & 2151.8 & 3095 &   & 2338 &   & --   &  \\
				28  & 2338 & 5    & 2351   & \textbf{2322}  & * & 2322.6 & 2569 &   & 2402 &   & --   &  \\
				29  & 3996 & 5    & 4192.6 & \textbf{3233}  &   & 3297   & 4539 &   & 3718 &   & --   &  \\
				30  & \textbf{4478} & 4    & 4544.75& 4736  &   & 4779.4 & 5904 &   & 4995 &   & --   &  \\ \midrule
				Lab & 3338 & 4    & 3366.75& \textbf{3334}  &   & 3400.6 & 5188 &   & 3444 &   & --   &  \\	\bottomrule
			\end{tabular}
			
			\label{tab:comparison}
\end{table*}

Table~\ref{tab:comparison} shows our next experiment, which is a comparison between Simulated Annealing, VLNS, Chuffed, the MIP solver CPLEX and the CP solver CP Optimizer. The time limit was set to 2 hours for each instance. CP Optimizer was run with 8 threads and the parameter \emph{FailureDirectedSearchEmphasis} was set to 4. For Simulated Annealing we used the implementation of \cite{Mischek2021AnnalsOfOR}, with the same parameter configuration.

It can easily be seen that CPLEX performed very poorly in comparison to the rest, although it should be noted that our model was developed with CP solvers in mind and thus might not be a perfect fit for MIP solvers.
Also, CPLEX was only run in single-threaded mode.

Chuffed on the contrary could find solutions for all instances and even found and proved optimal solutions for one more instance with the longer time-limit (instance 21, cf. \cite{Geibinger2019CPAIOR}).

The results for CP Optimizer are particularly interesting. 
While it only manages to prove the optimality of the 4 smallest instances, it generally finds better solutions for instances which cannot be closed by Chuffed.
The fact that CP Optimizer has trouble proving optimality is the main reason why we chose Chuffed for VLNS instead of CP Optimizer.

Simulated Annealing managed to find feasible solutions in all runs but two and most of the time the found solutions are better than the ones found by Chuffed, and competitive to those found by CP Optimizer.

As for VLNS, the search always finds feasible solutions and for 29 of the instances it found a solution with the best known objective.
Further, it could prove optimality using the custom lower bounds for 16 of the instances, including two large instances with 60 projects.

Results for the real-world instance can be compared to the reference schedule achieved via manual scheduling.
Firstly, this reference schedule omits 90 required resource assignments (88 employee and 2 equipment assignments), and thus does not fulfill all constraints.
Disregarding that, it achieves a penalty of 4074 under the same soft constraint weights as for the generated instances.
All solvers (except CPLEX and one of the runs of SA) found a feasible solution for this instance in the given time limit.
Both CPO, VLNS and SA (in 4 of 5 runs) even managed to find solutions with a markedly lower penalty than the reference schedule.
For VLNS, we could show that the best solution found is within 5.97\% of the optimum.

\begin{figure*}[ht!]
\begin{center}
    \includegraphics[width=\textwidth]{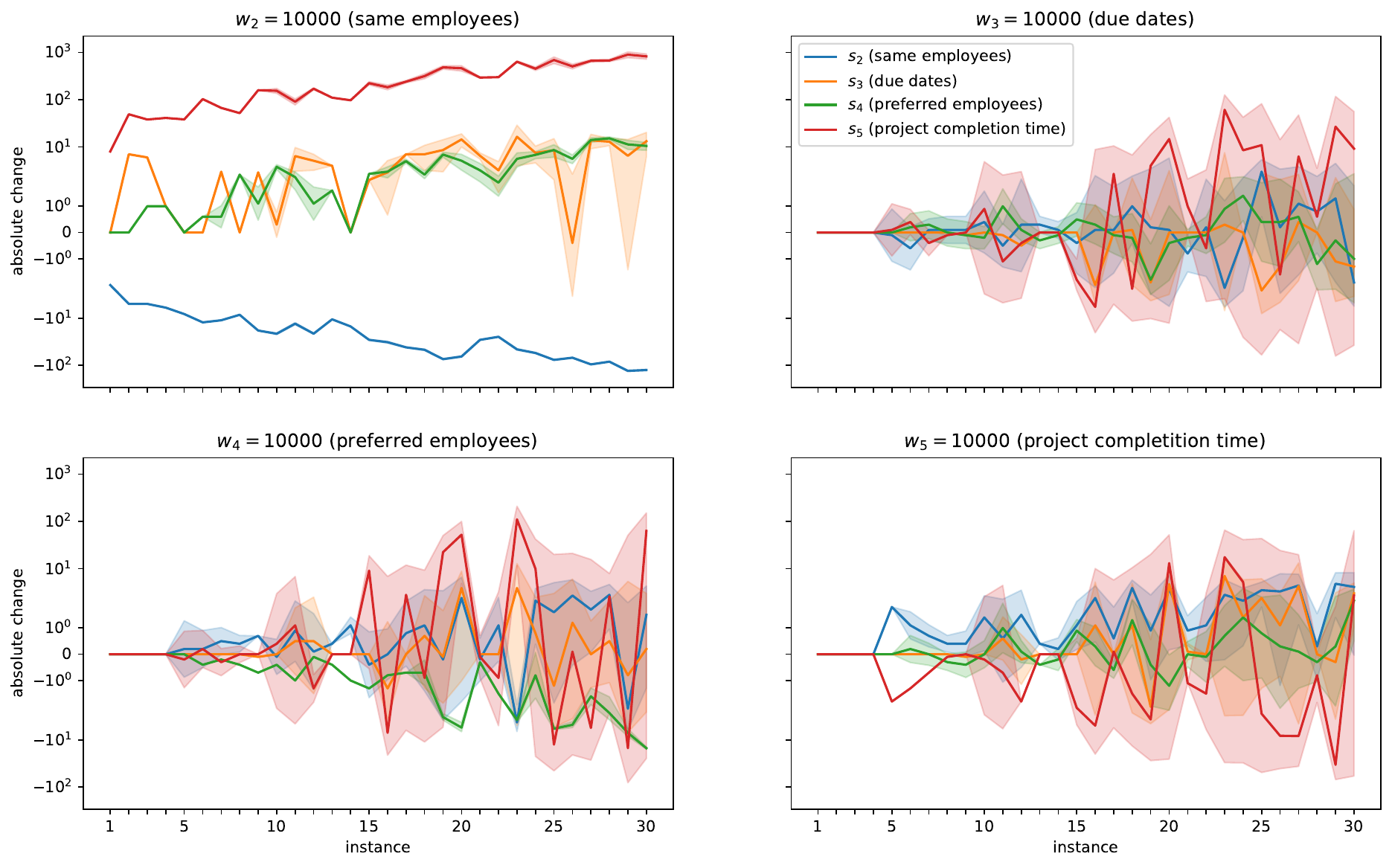}
\end{center}
\caption{Changes in penalties when prioritizing a single soft constraint.}
\label{fig:softconst}
\end{figure*}
\subsection{Soft Constraints Weights}
\label{sec:exp_softconst}

To further evaluate the effect of soft constraint weights, we performed an additional experiment. Starting from a baseline where each objective had a uniform weight of 1, we increased the weight of a single objective to 10000. We did so for each soft constraint and compared the results with the baseline scenario where all weights were set to 1.

Figure~\ref{fig:softconst} shows the absolute changes in the penalties $s_{i}$ $(2 \leq i \leq 5)$ when we prioritize a single objective in VLNS. For each instance, we ran VLNS ten times using a run time limit of 30 minutes. The lines represent the mean of the ten runs, whereas the shadows show the standard deviation. 

Setting VLNS to focus on minimizing the number of different employees (shown on the top left in the figure) seems to negatively affect all other soft constraints, especially the project completion times. This is hardly surprising, since minimizing the number of different employees usually involves executing jobs in a serial fashion, thus prolonging project durations. The worsening of the due date violations can be described similarly. For the rise in preferred employee violations, an explanation might be that minimizing the number different employees in a project sometimes means choosing an employee that is not preferred for the job.

For the other soft constraints, the results are less clear. In general, the total project completion time varies significantly between different runs even when the respective soft constraint is given high weight. Furthermore, the impacts the prioritization of a single soft constraints on the other penalties, might also depend on the particular instance but the high variance among the runs makes the drawing of any conclusions difficult.

\section{Conclusion}
In this paper we have investigated different possibilities to model a complex real-world project scheduling problem. For some of the constraints, we first experimented with approaches which were already used in related project scheduling problems. To deal with this more complex problem and larger instances we introduced several extensions in modeling. Utilising this exact method, we then implemented a VLNS meta-heuristic. We have evaluated our exact approach as well as the VLNS on a set of 30 benchmark instances. Using CP techniques we could find feasible solutions for all considered instances and 15 optimal solutions. Furthermore, the VLNS achieved even better results for most instances and managed to provide 6 provably optimal solutions which could not be found using CP.
VLNS also outperformed a local search procedure based on Simulated Annealing and found new best known solutions for several instances, including a real-world instance.

For the future, we plan to investigate exact and hybrid techniques to solve both stages of the TLSP including grouping and scheduling simultaneously.

\begin{acknowledgements}
The financial support by the Austrian Federal Ministry  of Labour  and  Economy,  the  National  Foundation  for  Research,  Technology  and  Development  and  the  Christian Doppler Research Association is gratefully acknowledged.
\end{acknowledgements}

%
%

\bibliographystyle{spbasic}      
\bibliography{references}   

\begin{thebibliography}{29}
\providecommand{\natexlab}[1]{#1}
\providecommand{\url}[1]{{#1}}
\providecommand{\urlprefix}{URL }
\expandafter\ifx\csname urlstyle\endcsname\relax
  \providecommand{\doi}[1]{DOI~\discretionary{}{}{}#1}\else
  \providecommand{\doi}{DOI~\discretionary{}{}{}\begingroup
  \urlstyle{rm}\Url}\fi
\providecommand{\eprint}[2][]{\url{#2}}

\bibitem[{Bartels and Zimmermann(2009)}]{Bartels2009DestructiveMode}
Bartels JH, Zimmermann J (2009) Scheduling tests in automotive {R}\&{D}
  projects. European Journal of Operational Research 193(3):805--819,
  \doi{10.1016/j.ejor.2007.11.010}

\bibitem[{Bellenguez and N{\'e}ron(2005)}]{Bellenguez2005MSPSP}
Bellenguez O, N{\'e}ron E (2005) Lower bounds for the multi-skill project
  scheduling problem with hierarchical levels of skills. In: Proceedings of the
  5th International Conference on the Practice and Theory of Automated
  Timetabling (PATAT 2005), Springer, LNCS, vol 3616, pp 229--243,
  \doi{10.1007/11593577\_14}

\bibitem[{Brucker et~al.(1999)Brucker, Drexl, M{\"o}hring, Neumann, and
  Pesch}]{Brucker1999Survey}
Brucker P, Drexl A, M{\"o}hring R, Neumann K, Pesch E (1999)
  Resource-constrained project scheduling: Notation, classification, models,
  and methods. European Journal of Operational Research 112(1):3--41,
  \doi{10.1016/S0377-2217(98)00204-5}

\bibitem[{Chu(2011)}]{Chu11}
Chu G (2011) Improving combinatorial optimization. PhD thesis, University of
  Melbourne, Australia, \urlprefix\url{http://hdl.handle.net/11343/36679}

\bibitem[{Dauz{\`e}re-P{\'e}r{\`e}s et~al.(1998)Dauz{\`e}re-P{\'e}r{\`e}s,
  Roux, and Lasserre}]{DauzerePeres1998}
Dauz{\`e}re-P{\'e}r{\`e}s S, Roux W, Lasserre J (1998) Multi-resource shop
  scheduling with resource flexibility. European Journal of Operational
  Research 107(2):289--305, \doi{10.1016/S0377-2217(97)00341-X}

\bibitem[{Demirovic et~al.(2018)Demirovic, Chu, and Stuckey}]{Demirovic18}
Demirovic E, Chu G, Stuckey PJ (2018) Solution-based phase saving for {CP:} {A}
  value-selection heuristic to simulate local search behavior in complete
  solvers. In: Proceedings of the 24th International Conference on Principles
  and Practice of Constraint Programming ({CP} 2018), Springer, LNCS, vol
  11008, pp 99--108, \doi{10.1007/978-3-319-98334-9_7}

\bibitem[{Drezet and Billaut(2008)}]{Drezet2008TimeConstraints}
Drezet LE, Billaut JC (2008) A project scheduling problem with labour
  constraints and time-dependent activities requirements. International Journal
  of Production Economics 112(1):217--225, \doi{10.1016/j.ijpe.2006.08.021}

\bibitem[{Elmaghraby(1977)}]{Elmaghraby1977activity}
Elmaghraby SE (1977) Activity networks: Project planning and control by network
  models. John Wiley \& Sons

\bibitem[{Feydy et~al.(2017)Feydy, Goldwaser, Schutt, Stuckey, and
  Young}]{Feydy17}
Feydy T, Goldwaser A, Schutt A, Stuckey PJ, Young KD (2017) Priority search
  with minizinc. In: Proceedings of ModRef 2017: The 16th International
  Workshop on Constraint Modelling and Reformulation at CP 2017,
  \urlprefix\url{https://ozgurakgun.github.io/ModRef2017/files/ModRef2017_PrioritySearchWithMiniZinc.pdf}

\bibitem[{Geibinger et~al.(2019)Geibinger, Mischek, and
  Musliu}]{Geibinger2019CPAIOR}
Geibinger T, Mischek F, Musliu N (2019) Investigating constraint programming
  for real world industrial test laboratory scheduling. In: Proceedings of the
  16th International Conference on the Integration of Constraint Programming,
  Artificial Intelligence, and Operations Research (CPAIOR 2019), Springer,
  LNCS, vol 11494, pp 304--319, \doi{10.1007/978-3-030-19212-9_20}

\bibitem[{Hartmann and Briskorn(2010)}]{Hartmann2010}
Hartmann S, Briskorn D (2010) A survey of variants and extensions of the
  resource-constrained project scheduling problem. European Journal of
  Operational Research 207(1):1--14, \doi{10.1016/j.ejor.2009.11.005}

\bibitem[{IBM and CPLEX(2017{\natexlab{a}})}]{cpoptimizer}
IBM, CPLEX (2017{\natexlab{a}}) 12.8.0 {IBM} {ILOG} {CPLEX} {O}ptimization
  {S}tudio {CP} {O}ptimizer user's manual.
  \url{https://www.ibm.com/analytics/cplex-cp-optimizer}

\bibitem[{IBM and CPLEX(2017{\natexlab{b}})}]{cplex}
IBM, CPLEX (2017{\natexlab{b}}) 12.8.0 {IBM} {ILOG} {CPLEX} {O}ptimization
  {S}tudio {CPLEX} user's manual.
  \url{https://www.ibm.com/analytics/cplex-optimizer}

\bibitem[{Kirkpatrick et~al.(1983)Kirkpatrick, Gelatt, and
  Vecchi}]{Kirkpatrick1983SimulatedAnnealing}
Kirkpatrick S, Gelatt CD, Vecchi MP (1983) Optimization by simulated annealing.
  Science 220(4598):671--680, \doi{10.1126/science.220.4598.671}

\bibitem[{Laborie et~al.(2018)Laborie, Rogerie, Shaw, and
  Vil\'{\i}m}]{Laborie18}
Laborie P, Rogerie J, Shaw P, Vil\'{\i}m P (2018) {IBM} {ILOG} {CP} {O}ptimizer
  for scheduling. Constraints 23(2):210–250, \doi{10.1007/s10601-018-9281-x}

\bibitem[{Mika et~al.(2015)Mika, Walig{\'o}ra, and Węglarz}]{Mika2015Survey}
Mika M, Walig{\'o}ra G, Węglarz J (2015) Overview and state of the art. In:
  Schwindt C, Zimmermann J (eds) Handbook on Project Management and Scheduling
  Vol.1, Springer, pp 445--490

\bibitem[{Mischek and Musliu(2018{\natexlab{a}})}]{MischekPatat18}
Mischek F, Musliu N (2018{\natexlab{a}}) A local search framework for
  industrial test laboratory scheduling. In: Proceedings of the 12th
  International Conference on the Practice and Theory of Automated Timetabling
  (PATAT 2018), pp 465--467,
  \urlprefix\url{https://patatconference.org/patat2018/files/proceedings/paper33.pdf}

\bibitem[{Mischek and Musliu(2018{\natexlab{b}})}]{Mischek2018TechReport}
Mischek F, Musliu N (2018{\natexlab{b}}) The test laboratory scheduling
  problem. Technical report, Christian Doppler Laboratory for Artificial
  Intelligence and Optimization for Planning and Scheduling, TU Wien, CD-TR
  2018/1,
  \urlprefix\url{https://www.dbai.tuwien.ac.at/staff/fmischek/TLSP/TLSP.pdf}

\bibitem[{Mischek and Musliu(2021)}]{Mischek2021AnnalsOfOR}
Mischek F, Musliu N (2021) A local search framework for industrial test
  laboratory scheduling. Annals of Operations Research 302:533--562,
  \doi{10.1007/s10479-021-04007-1}

\bibitem[{Nethercote et~al.(2007)Nethercote, Stuckey, Becket, Brand, Duck, and
  Tack}]{Nethercote07}
Nethercote N, Stuckey PJ, Becket R, Brand S, Duck GJ, Tack G (2007) Minizinc:
  Towards a standard {CP} modelling language. In: Proceedings of the 13th
  International Conference on Principles and Practice of Constraint Programming
  ({CP} 2007), Springer, LNCS, vol 4741, pp 529--543,
  \doi{10.1007/978-3-540-74970-7_38}

\bibitem[{Nudtasomboon and Randhawa(1997)}]{Nudtasomboon1997}
Nudtasomboon N, Randhawa SU (1997) Resource-constrained project scheduling with
  renewable and non-renewable resources and time-resource tradeoffs. Computers
  \& Industrial Engineering 32(1):227--242, \doi{10.1016/S0360-8352(96)00212-4}

\bibitem[{Palpant et~al.(2004)Palpant, Artigues, and Michelon}]{Palpant2004}
Palpant M, Artigues C, Michelon P (2004) Lssper: Solving the
  resource-constrained project scheduling problem with large neighbourhood
  search. Annals of Operations Research 131(1):237--257,
  \doi{10.1023/B:ANOR.0000039521.26237.62}

\bibitem[{Pritsker et~al.(1969)Pritsker, Waiters, and Wolfe}]{Pritsker1969}
Pritsker AAB, Waiters LJ, Wolfe PM (1969) Multiproject scheduling with limited
  resources: A zero-one programming approach. Management Science 16(1):93--108,
  \doi{10.1287/mnsc.16.1.93}

\bibitem[{Salewski et~al.(1997)Salewski, Schirmer, and
  Drexl}]{Salewski1997ModeIdentity}
Salewski F, Schirmer A, Drexl A (1997) Project scheduling under resource and
  mode identity constraints: Model, complexity, methods, and application.
  European Journal of Operational Research 102(1):88--110,
  \doi{10.1016/S0377-2217(96)00219-6}

\bibitem[{Schulte et~al.(2018)Schulte, Lagerkvist, and Tack}]{gecode}
Schulte C, Lagerkvist M, Tack G (2018) Gecode 6.10 reference documentation.
  \url{https://www.gecode.org}

\bibitem[{Schwindt and Trautmann(2000)}]{Schwindt2000Batch}
Schwindt C, Trautmann N (2000) Batch scheduling in process industries: an
  application of resource-constrained project scheduling. OR-Spektrum
  22(4):501--524, \doi{10.1007/s002910000042}

\bibitem[{Szeredi and Schutt(2016)}]{Szeredi16}
Szeredi R, Schutt A (2016) Modelling and solving multi-mode
  resource-constrained project scheduling. In: Proceedings of the 22nd
  International Conference on Principles and Practice of Constraint Programming
  ({CP} 2016), Springer, LNCS, vol 9892, pp 483--492,
  \doi{10.1007/978-3-319-44953-1_31}

\bibitem[{Węglarz et~al.(2011)Węglarz, J\'{o}zefowska, Mika, and
  Walig\'{o}ra}]{Weglarz2011MRCPSPSurvey}
Węglarz J, J\'{o}zefowska J, Mika M, Walig\'{o}ra G (2011) Project scheduling
  with finite or infinite number of activity processing modes – a survey.
  European Journal of Operational Research 208(3):177--205,
  \doi{10.1016/j.ejor.2010.03.037}

\bibitem[{Young et~al.(2017)Young, Feydy, and Schutt}]{Young17}
Young KD, Feydy T, Schutt A (2017) Constraint programming applied to the
  multi-skill project scheduling problem. In: Proceedings of the 23rd
  International Conference on Principles and Practice of Constraint Programming
  ({CP} 2017), LNCS, vol 10416, pp 308--317, \doi{10.1007/978-3-319-66158-2_20}

\end{thebibliography}

\end{document}